\pdfoutput=1
\documentclass{llncs}

\usepackage[width=122mm,left=30.7mm,paperwidth=183.5mm,height=193mm,top=22.25mm,paperheight=237.5mm]{geometry}

\usepackage{graphicx}
\usepackage{amsmath,amssymb}
\usepackage{color}
\usepackage{xcolor,colortbl}
\usepackage[colorlinks=true]{hyperref}

\definecolor{tblcolor}{gray}{0.75}

\newcommand{\Co}{\mathbb{C}}
\newcommand{\pp}{{p^+}}

\newcommand{\rgt}{r}

\newcommand{\etal}{\emph{et al.}}
\newcommand{\ie}{\emph{i.e.},~}
\newcommand{\eg}{\emph{e.g.},~}
\renewcommand{\paragraph}{\subsubsection}

\begin{document}
\pagestyle{plain}
\mainmatter

\title{A Neural Approach to Blind Motion Deblurring}
\author{Ayan Chakrabarti}
\institute{Toyota Technological Institute at Chicago\\\email{ayanc@ttic.edu}}

\maketitle
\thispagestyle{plain}

\begin{abstract}
We present a new method for blind motion deblurring that uses a neural network trained to compute estimates of sharp image patches from observations that are blurred by an unknown motion kernel. Instead of regressing directly to patch intensities, this network learns to predict the complex Fourier coefficients of a deconvolution filter to be applied to the input patch for restoration. For inference, we apply the network independently to all overlapping patches in the observed image, and average its outputs to form an initial estimate of the sharp image. We then explicitly estimate a single global blur kernel by relating this estimate to the observed image, and finally perform non-blind deconvolution with this kernel. Our method exhibits accuracy and robustness close to state-of-the-art iterative methods, while being much faster when parallelized on GPU hardware.

\end{abstract}

\section{Introduction}
\label{sec:intro}

Photographs captured with long exposure times using hand-held cameras are often degraded by blur due to camera shake. The ability to reverse this degradation and recover a sharp image is attractive to photographers, since it allows rescuing an otherwise acceptable photograph. Moreover, if this ability is \emph{consistent} and can be relied upon post-acquisition, it gives photographers more flexibility at the time of capture, for example, in terms of shooting with a zoom-lens without a tripod, or trading off exposure time with ISO in low-light settings. Beginning with the seminal work of Fergus \etal~\cite{fergus06}, the last decade has seen considerable progress~\cite{irani,sun2013,xujia,cholee,levinetal,krishnan,choetal} in the development of effective blind motion deblurring methods that seek to estimate camera motion in terms of the induced blur kernel, and then reverse its effect. This progress has been helped by the development of principled evaluation on standard benchmarks~\cite{sun2013,levin2009}, that measure performance over a large and diverse set of images.

Some deblurring algorithms~\cite{xujia,cholee} emphasize efficiency, and use inexpensive processing of image features to quickly estimate the motion kernel. Despite their speed, these methods can yield remarkably accurate kernel estimates and achieve high-quality restoration for many images, making them a practically useful post-processing tool for photographers. However, due to their reliance on relatively simple heuristics, they also have poor outlier performance and can fail on a significant fraction of blurred images. Other methods are iterative---they reason with parametric prior models for natural images and motion kernels, and use these priors to successively improve the algorithm's estimate of the sharp image and the motion kernel. The two most successful deblurring algorithms fall~\cite{irani,sun2013} in this category, and while they are able to outperform previous methods by a significant margin, they also have orders of magnitude longer running times.

In this work, we explore whether discriminatively trained neural networks can match the performance of traditional methods that use generative natural image priors, and do so without multiple iterative refinements. Our work is motivated by recent successes in the use of neural networks for other image restoration tasks (\eg \cite{ndnz,ndnz2,ndeblur1,ndeblur2}). This includes methods~\cite{ndeblur1,ndeblur2} for \emph{non-blind} deconvolution, \ie restoring a blurred image when the blur kernel is known. While the estimation problem in blind deconvolution is significantly more ill-posed than the non-blind case, these works provide insight into the design process of neural architectures for deconvolution. 

\begin{figure}[!t]
  \centering
  \includegraphics[width=\textwidth]{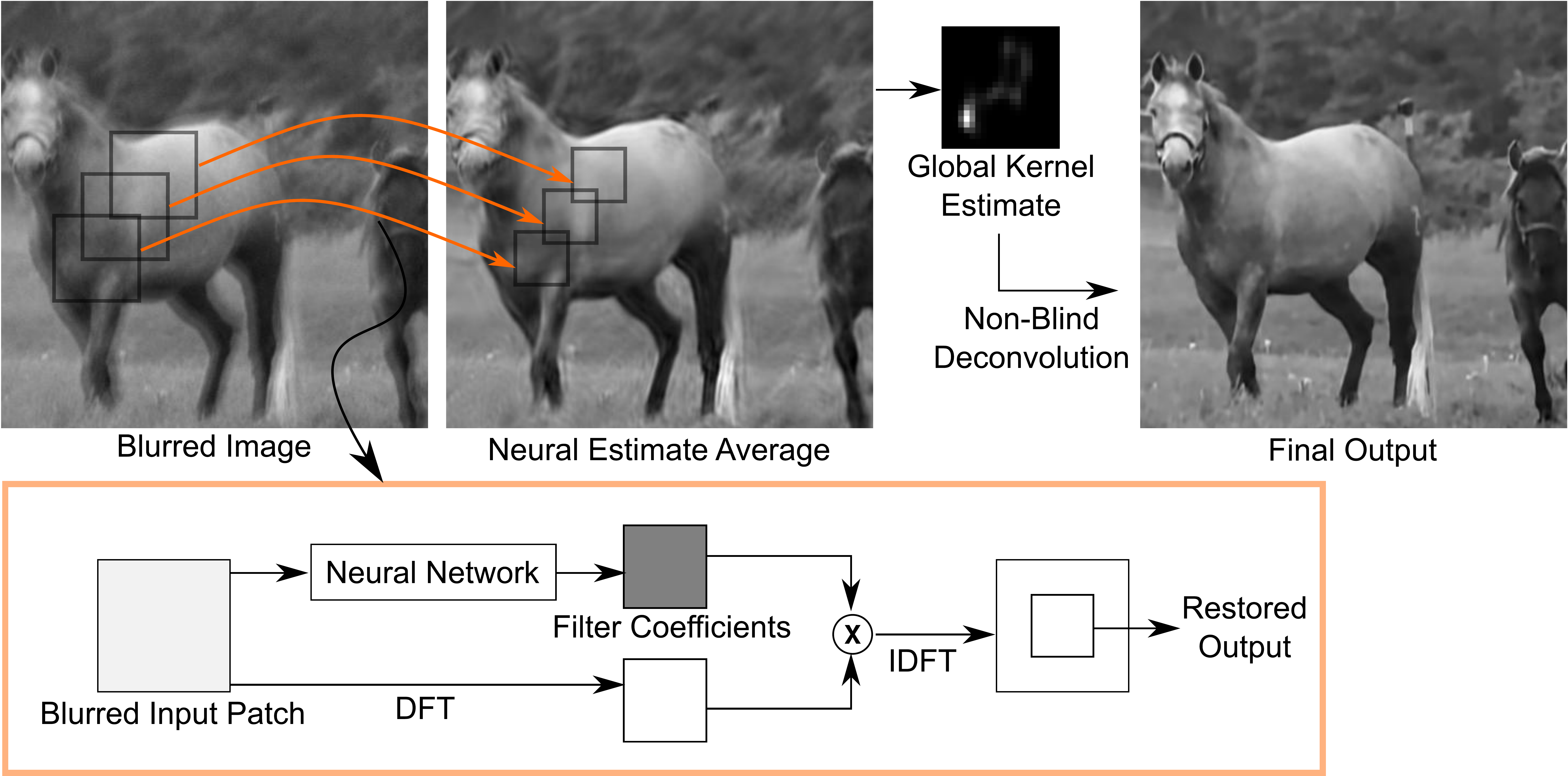}
  \caption{Neural Blind Deconvolution. Our method uses a neural network trained for per-patch blind deconvolution. Given an input patch blurred by an unknown motion blur kernel, this network predicts the Fourier coefficients of a filter to be applied to that input for restoration. For inference, we apply this network independently on all overlapping patches in the input image, and compose their outputs to form an initial estimate of the sharp image. We then infer a single global blur kernel that relates the input to this initial estimate, and use that kernel for non-blind deconvolution.}
  \label{fig:system}
\end{figure}

Hradi{\v{s}} \etal~\cite{textdb} explored the use of neural networks for blind deconvolution on images of text. Since text images are highly structured---two-tone with thin sparse contours---a standard feed-forward architecture was able to achieve successful restoration. Meanwhile, Sun \etal~\cite{mbclass} considered a version of the problem with restrictions on motion blur types, and were able to successfully train a neural network to identify the blur in an observed natural image patch from among a small discrete set of oriented box blur kernels of various lengths. Recently, Schuler \etal~\cite{l2d} tackled the general blind motion deblurring problem using a neural architecture designed to mimic the computational steps of traditional iterative deblurring methods. They designed learnable layers to carry out extraction of salient local image features and kernel estimation based on these features, and stacked multiple copies of these layers to enable iterative refinement. Remarkably, they were able to train this multi-stage network with relative success. However, while their initial results are very encouraging, their current performance still significantly lags behind the state of the art~\cite{irani,sun2013}---especially when the unknown blur kernel is large.

In this paper, we propose a new approach for blind deconvolution of natural images degraded by arbitrary motion blur kernels due to camera shake. At the core of our algorithm is a neural network trained to restore individual image patches. This network differs from previous architectures in two significant ways:
\begin{enumerate}
\item Rather than formulate the prediction task as blur kernel estimation through iterative refinement (as in \cite{l2d}), or as direct regression to deblurred intensity values (as in \cite{ndnz,ndnz2,ndeblur1,ndeblur2}), we train our network to output the complex Fourier coefficients of a \emph{deconvolution} filter to be applied to the input patch.

\item We use a multi-resolution frequency decomposition to encode the input patch, and limit the connectivity of initial network layers based on locality in frequency (analogous to convolutional layers that are limited by locality in space). This leads to a significant reduction in the number of weights to be learned during training, which proves crucial since it allows us to successfully train a network that operates on large patches, and therefore can reason about large blur kernels (\eg in comparison to \cite{l2d}).
\end{enumerate}

For whole image restoration, the network is independently applied to every overlapping patch in the input image, and its outputs are composed to form an initial estimate of the latent sharp image. Despite reasoning with patches independently and not sharing information about a common global motion kernel, we find that this procedure by itself performs surprisingly well. We show that these results can be further improved by using the restored image to compute a global blur kernel estimate, which is finally used for non-blind deconvolution.  Evaluation on a standard benchmark~\cite{sun2013} demonstrates that our approach is competitive when considering accuracy, robustness, and running time.

\section{Patch-wise Neural Deconvolution}
\label{sec:neural}

Let $y[n]$ be the observed image of a scene blurred due to camera motion, and $x[n]$ the corresponding latent sharp image that we wish to estimate, with $n\in\mathbb{Z}^2$ indexing pixel location. The degradation due to blur can be approximately modeled as convolution with an unknown blur kernel $k$:
\begin{equation}
  \label{eq:obs}
  y[n] = (x * k)[n] + \epsilon[n],\quad k[n] \geq 0, \sum_n k[n] = 1,
\end{equation}
where $*$ denotes convolution, and $\epsilon[n]$ is i.i.d.~Gaussian noise. 

As shown in Fig.~\ref{fig:system}, the central component of our algorithm is a neural network that carries out restoration locally on individual patches in $y[n]$. Formally, our goal is to design a network that is able to recover the sharp intensity values $x_p=\{x[n]:n\in p\}$ of a patch $p$, given as input a larger patch $y_\pp=\{x[n]:n\in p^+\}$, $\pp\supset p$ from the observed image. The larger input is necessary since values in $x_p[n]$, especially near the boundaries of $p$, can depend on those outside $y_p[n]$. In practice, we choose $\pp$ to be of size $65\times 65$, with its central $33\times 33$ patch corresponding to $p$. In this section, we describe our formulation of the prediction task for this network, its architecture and connectivity, and our approach to training it.

\subsection{Restoration by Predicting Deconvolution Filter Coefficients}
\label{sec:form}

As depicted in Fig.~\ref{fig:system}, the output of our network are the complex discrete Fourier transform (DFT) coefficients $G_\pp[z] \in\Co$ of a deconvolution filter, where $z$ indexes two-dimensional spatial frequencies in the DFT. This filter is then applied DFT $Y_\pp[z]$ of the input patch $y_\pp[n]$:
\begin{equation}
  \label{eq:cflt}
  \hat{X}_\pp[z] = G_\pp[z]~\times~Y_\pp[z].
\end{equation}
Our estimate $\hat{x}_p[n]$ of the sharp image patch is computed by taking the inverse discrete Fourier transform (IDFT) of $\hat{X}_\pp[z]$, and then cropping out the central patch $p \subset \pp$. Since $x[n]$ and $y[n]$ are both real valued and $k$ is unit sum, we assume that $G_\pp[z] = G_\pp^*[-z]$, and $G_\pp[0] = 1$. Therefore, the network only needs to output $(|\pp|-1)/2$ unique complex numbers to characterize $G_\pp$, where $|\pp|$ is the number of pixels in $\pp$.

Our training objective is that the output coefficients $G_\pp[z]$ be optimal with respect to the quality of the final sharp intensities $\hat{x}_p[n]$. Specifically, the loss function for the network is defined as the mean square error (MSE) between the predicted and true sharp intensity values $\hat{x}_p[n]$ and $x_p[n]$:
\begin{equation}
  \label{eq:loss}
  L(\hat{x}_p,x_p) = \frac{1}{|p|}\sum_{n\in p} (\hat{x}_p[n]-x_p[n])^2.
\end{equation}
Note both the IDFT and the filtering in \eqref{eq:cflt} are linear operations, and therefore it is trivial to back-propagate the gradients of \eqref{eq:loss} to the outputs $G_\pp[z]$, and subsequently to all layers within the network.

\paragraph{Motivation.} As with any neural-network based method, the validation of the design choices in our approach ultimately has to be empirical. However, we attempt to provide the reader with some insight into our motivation for making these choices. We begin by considering the differences between predicting deconvolution filter coefficients and regressing directly to pixel intensities $x_p[n]$, as was done in most prior neural restoration methods~\cite{ndnz,ndnz2,ndeblur1,ndeblur2}. Indeed, since we use the predicted coefficients to estimate $x_p[n]$ and define our loss with respect to the latter, our overall formulation \emph{can} be interpreted as a regression to $x_p[n]$. However, our approach enforces a specific parametric form being enforced on the learned mapping from $y_\pp[n]$ to $x_p[n]$. In other words, the notion that the sharp and blurred image patches are related by convolution is ``baked-in'' to the network's architecture. Additionally, providing $Y_\pp[z]$ separately at the output alleviates the need for the layers within our network to retain a linear encoding of the input patch all the way to the output.

Another alternative formulation could have been to set-up the network to predict the blur kernel $k$ itself like in \cite{l2d}, which also encodes the convolutional relationship between the network's input and output. However, remember that our network works on local patches independently. For many patches, inferring the blur kernel may be impossible from local information alone---for example, a patch with only a vertical edge would have no content in horizontal frequencies, making it impossible to infer the horizontal structure of the kernel. But in these cases, it would still be possible to compute an optimal deconvolution filter and restored image patch (in our example, the horizontal frequency values of $G_\pp[z]$ would not matter). Moreover, our goal is to recover the restored image patch and estimating the kernel solves only a part of the problem, since non-blind deconvolution is not trivial. In contrast, our predicted deconvolution filter can be directly applied for restoration, and because it is trained with respect to restoration quality, the network learns to generate these predictions by reasoning both about the unknown kernel and sharp image content.

It may be helpful to consider what the optimal values of $G_\pp[z]$ should be. One interpretation for these values can be derived from Wiener deconvolution~\cite{wiener}, in which ideal restoration is achieved by applying a filter using \eqref{eq:cflt} with coefficients given by
\begin{equation}
  \label{eq:wnr}
G_\pp[z] = \left(\left|K[z]\right|^2S_\pp[z]+\sigma_\epsilon^2\right)^{-1} K^*[z]S_\pp[z].
\end{equation}
Here, $K[z]$ is the DFT of the kernel $k$, and $S_\pp[z]$ is the spectral profile of $x_\pp[n]$ (\ie a DFT of its auto-correlation function). Note that in blind deconvolution, both $S_\pp[z]$ and $K[z]$ are unknown and iterative algorithms can be interpreted as explicitly estimating these quantities through sequential refinement. In contrast, our network is discriminatively trained to directly predict the ratio in \eqref{eq:wnr}.

\begin{figure}[!t]
  \centering
  \includegraphics[width=0.9\textwidth]{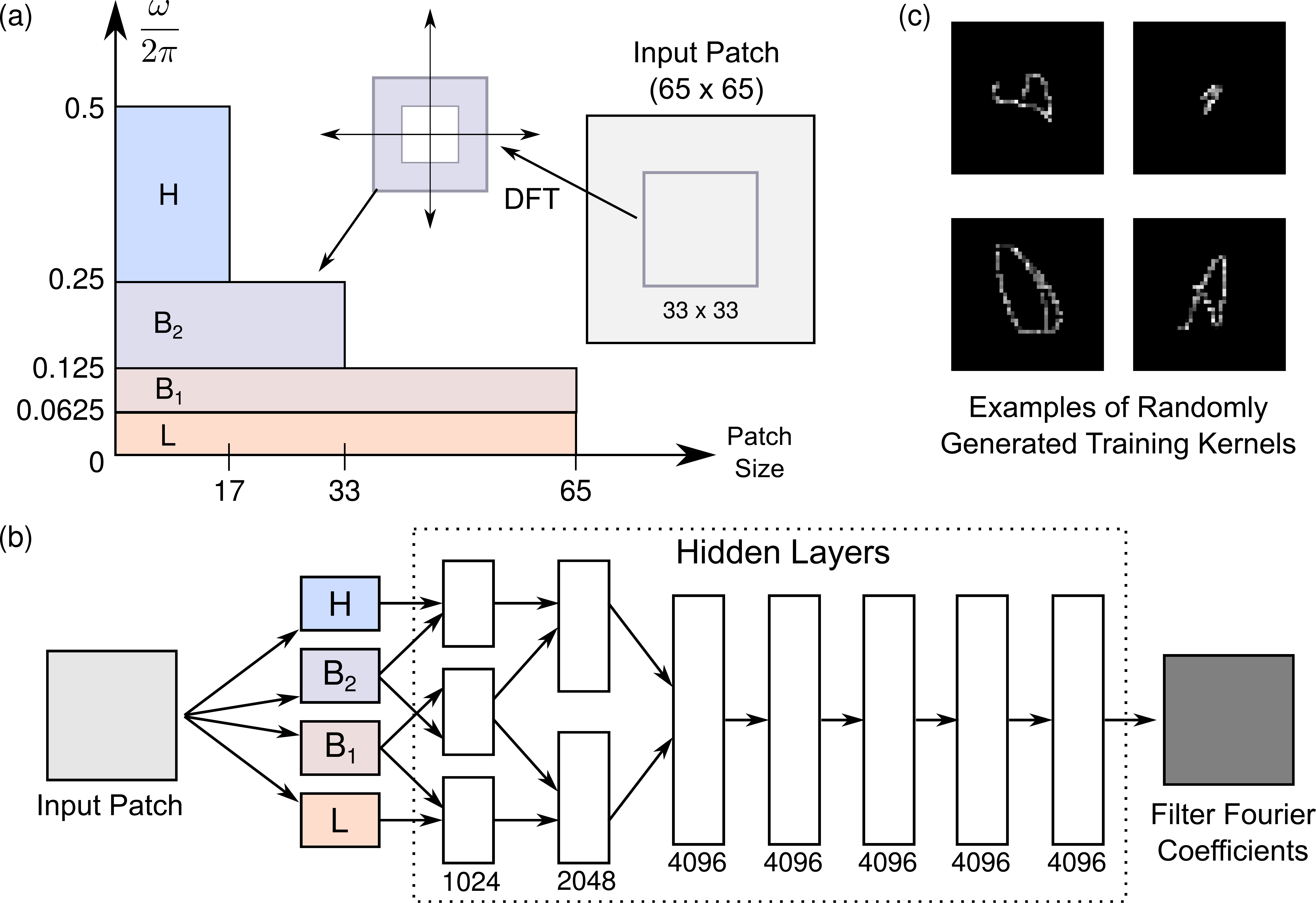}
  \caption{Network Architecture. (a) To limit the number of weights in the network, we use a multi-resolution decomposition to encode the input patch into four ``bands'', containing low-pass (L), band-pass ($B_1$, $B_2$), and high-pass $H$ frequency components. Higher frequencies are sampled at a coarser resolution, and computed from a smaller patch centered on the input. (b) Our network regresses from this input encoding to the complex Fourier coefficients of a restoration filter, and contains seven hidden layers (each followed by a ReLU activation). The connectivity of the initial layers is limited to adjacent frequency bands. (c) Our network is trained with randomly generated synthetic motion blur kernels of different sizes.}
  \label{fig:arch}
\end{figure}

\subsection{Network Architecture}

Our network needs to work with large input patches in order to successfully handle large blur kernels. This presents a challenge in terms of the number of weights to be learned and the feasibility of training since, as observed by \cite{ndeblur2} and by us in our own experiments, the traditional strategy of making the initial layers convolutional with limited support performs poorly for deconvolution. This why the networks for blind deconvolution in \cite{ndeblur1,ndeblur2} have either used only fully connected layers~\cite{ndeblur1}, or large oriented on-dimensional convolutional layers~\cite{ndeblur2}.

We adopt a novel approach to parameterizing the input patch and defining the connectivity of the initial layers in our network. Specifically, we use a multi-resolution decomposition strategy (illustrated in Fig.~\ref{fig:arch}~(a)) where higher spatial frequencies are sampled with lower resolution. We compute DFTs at three different levels, corresponding to patches of three different sizes ($17\times 17, 33\times 33,$ and $65\times 65$) centered on the input patch, and from each retain the coefficients corresponding to $4 < \max |z| \leq 8$. Here, $\max |z|$ represents the larger magnitude of the two components (horizontal and vertical) of the frequency indices in $z$. 

This decomposition gives us $104$ independent complex coefficients (or $208$ scalars) from each DFT level that we group into ``bands''.  Note that the indices $z$ correspond to different spatial frequencies $\omega$ for different sized DFTs, with coefficients from the smaller-size DFTs representing a coarser sampling in the frequency domain. Therefore, the three bands above  correspond to high- and band-pass components of the input patch. We also construct a low-pass band by including the  coefficients corresponding to $\max |z| \leq 4$ from the largest (\ie, $65\times 65$) decomposition. This band only has $81$ scalar components ($40$ complex coefficients and a scalar DC coefficient). As suggested in \cite{lecun-98b}, we apply a de-correlating linear transform to the coefficients of each band, based on their empirical covariance on input patches in the training set.

Note that our decomposition also entails a dimensionality reduction---the total number of coefficients in the four bands is lower than the size of the input patch. Such a reduction may have been problematic if the network were directly regressing to patch intensities. However, we find this approximate representation suffices for our task of predicting filter coefficients, since the full input patch $y_\pp[n]$ (in the form of its DFT) is separately provided to \eqref{eq:cflt} for the computation of the final output $\hat{x}_p[n]$.

As depicted in Fig.~\ref{fig:arch}~(b), we use a feed-forward network architecture with seven hidden layers to predict the coefficients $G_\pp[z]$ from our encoding of the observed blurry input patch. Units in the first layer are only connected to input coefficients from pairs of adjacent frequency bands---with groups of $1024$ units connected to each pair. Note that these groups do not share weights. We adopt a similar strategy for the next layer, connecting units to pairs of adjacent groups from the first layer. Each group in this layer has $2048$ units. Restricting connectivity in this way, based on locality in frequency, reduces the number of weights in our network, while still allowing good prediction in practice. This is not entirely surprising, since many iterative algorithms (including \cite{irani,sun2013}) also divide the inference task into sequential coarse-to-fine reasoning at individual scales. All remaining layers in our network are fully connected with $4096$ units each. Units in all hidden layers have ReLU activations~\cite{relu}.

\subsection{Training}
\label{sec:train}

Our network was trained on a synthetic dataset that is entirely disjoint from the evaluation benchmark~\cite{sun2013}. This was constructed by extracting sharp image patches from images in the Pascal VOC 2012 dataset~\cite{voc}, blurring them with synthetically generated kernels, and adding Gaussian noise. We set the noise standard deviation to $1\%$ to match the noise level in the benchmark~\cite{sun2013}. 

The synthetic motion kernels were generated by randomly sampling six points in a limited size grid (we generate an equal number of kernels from grid sizes of $8\times 8$, $16\times 16$, and $24\times 24$), fitting a spline through these points, and setting the kernel values at each pixel on this spline to a value sampled from a Gaussian distribution with mean one and standard deviation of half. We then clipped these values to be positive, and normalized the kernel to be unit sum. 

There is an inherent phase ambiguity in blind deconvolution---one can apply equal but opposite translations to the blur kernel and sharp image estimates to come up with equally plausible explanations for an observation. While this ambiguity need not be resolved globally, we need our local $G_\pp[z]$ estimates in overlapping patches to have consistent phase. Therefore, we ensured that the training kernels have a ``canonical'' translation by centering them so that each kernel's center of mass (weighted by kernel values) is at the center of the window. Figure~\ref{fig:arch}~(c) shows some of the kernels generated using this approach.

We constructed separate training and validation sets with different sharp patches and randomly generated kernels. We used about 520,000 and 3,000 image patches and 100,000 and 3,000 kernels for the training and validation sets respectively. While we extracted multiple patches from the same image, we ensured that the training and validation patches were drawn from different images. To minimize disk access, we loaded the entire set of sharp patches and kernels into memory. Training data was generated on the fly by selecting random pairs of patches and kernels, and convolving the two to create the input patch. We also used rotated and mirrored versions of the sharp patches. This gave us a near inexhaustible supply of training data. Validation data was also generated on the fly, but we always chose the same pairs of patches and kernels to ensure that validation error could be compared across iterations.

We used stochastic gradient descent for minimizing the loss function \eqref{eq:loss}, with  a batch-size of 512 and a momentum value of 0.9. We trained the network for a total of 1.8 million iterations, which took about 3 days using an NVIDIA Titan X GPU. We used a learning rate of 32 (higher rates caused gradients to explode) for the first 800k iterations, at which point validation error began to plateau. For the remaining iterations, we dropped the rate by a factor of $\sqrt{2}$ every 100k iterations. We kept track of the validation error across iterations, and at the end of training, used the weights that yielded the lowest value of that error. 

\section{Whole Image Restoration}
\label{sec:rest}

Given an observed blurry image $y[n]$, we consider all overlapping patches $y_\pp$ in the image, and use our trained network to compute estimates $\hat{x}_p$ of their latent sharp versions. We then combines these restored patches to form an initial an estimate $x_N[n]$ of the sharp image, by setting $x_N[n]$ to the average of its estimates $\hat{x}_p[n]$ from all patches $p \ni n$ that contain it, using a Hanning window to weight the contributions from different patches.

While this feed-forward and purely local procedure achieves reasonable restoration, we have so far not taken into account the fact that the entire image has been blurred by the same motion kernel. To do so, we compute an estimate of the global kernel $k[n]$, by relating the observed image $y[n]$ to our neural-average estimate $x_N[n]$. Formally, we estimate this kernel $k[n]$ as
\begin{equation}
  \label{eq:kest1}
  k = \arg \min \sum_i \|(k * (f_i * x_N)) - (f_i * y) \|^2,
\end{equation}
subject to the constraint that $k[n] > 0$ and $\sum_n k[n] = 1$. We do not assume that the size of the kernel is known, and always estimate $k[n]$ within a fixed-size support ($51\times 51$ as is standard for the benchmark~\cite{sun2013}). Here, $f_i[n]$ are various derivative filters (we use first and second order derivatives at 8 orientations). Like in \cite{sun2013}, we only let strong gradients participate in the estimation process by setting values of $(f_i * x_N)$ to zero except those at the two percent pixel locations with the highest magnitudes.

This approach to estimating a global kernel from an estimate of the latent sharp image is fairly standard. But while it is typically used repeatedly within an iterative procedure that refines the estimates of the sharp image as well (\eg in \cite{irani,sun2013}), we estimate the kernel only once from the neural average output. 

We adopt a relatively simple and fast approach to optimizing \eqref{eq:kest1} under the positivity and unit sum constraints on $k$. Specifically, we minimize $L1$ regularized versions of the objective:
\begin{equation}
  \label{eq:kest2}
  k_\lambda = \arg \min \sum_i \|(k * (f_i * x_N)) - (f_i * y)  \|^2 + \lambda \sum_n |k[n]|,
\end{equation}
for a small range of values for the regularization weight $\lambda$. This optimization, for each value of $\lambda$, can be done very efficiently in the Fourier domain using half-quadratic splitting~\cite{hsq}. We clip each kernel estimate $k_\lambda[n]$ to be positive, set very small or isolated values to zero, and normalize the result to be unit sum. We then pick the kernel $k_\lambda[n]$ which yields the lowest value of the original un-regularized cost in \eqref{eq:kest1}. Given this estimate of the global kernel, we use EPLL~\cite{epll}---a state-of-the-art \emph{non-blind} deconvolution algorithm---to deconvolve $y[n]$ and arrive at our final estimate of the sharp image $x[n]$.

\section{Experiments}
\label{sec:exp}

We evaluate our approach on the benchmark dataset of Sun \etal~\cite{sun2013}, which consists of 640 blurred images generated from 80 high quality natural images, and 8 real motion blur kernels acquired by Levin \etal~\cite{levin2009}. We begin by analyzing patch-wise predictions from our neural network, and then compare the performance of our overall algorithm to the state of the art.

\begin{figure}[!t]
  \centering
  \includegraphics[width=\textwidth]{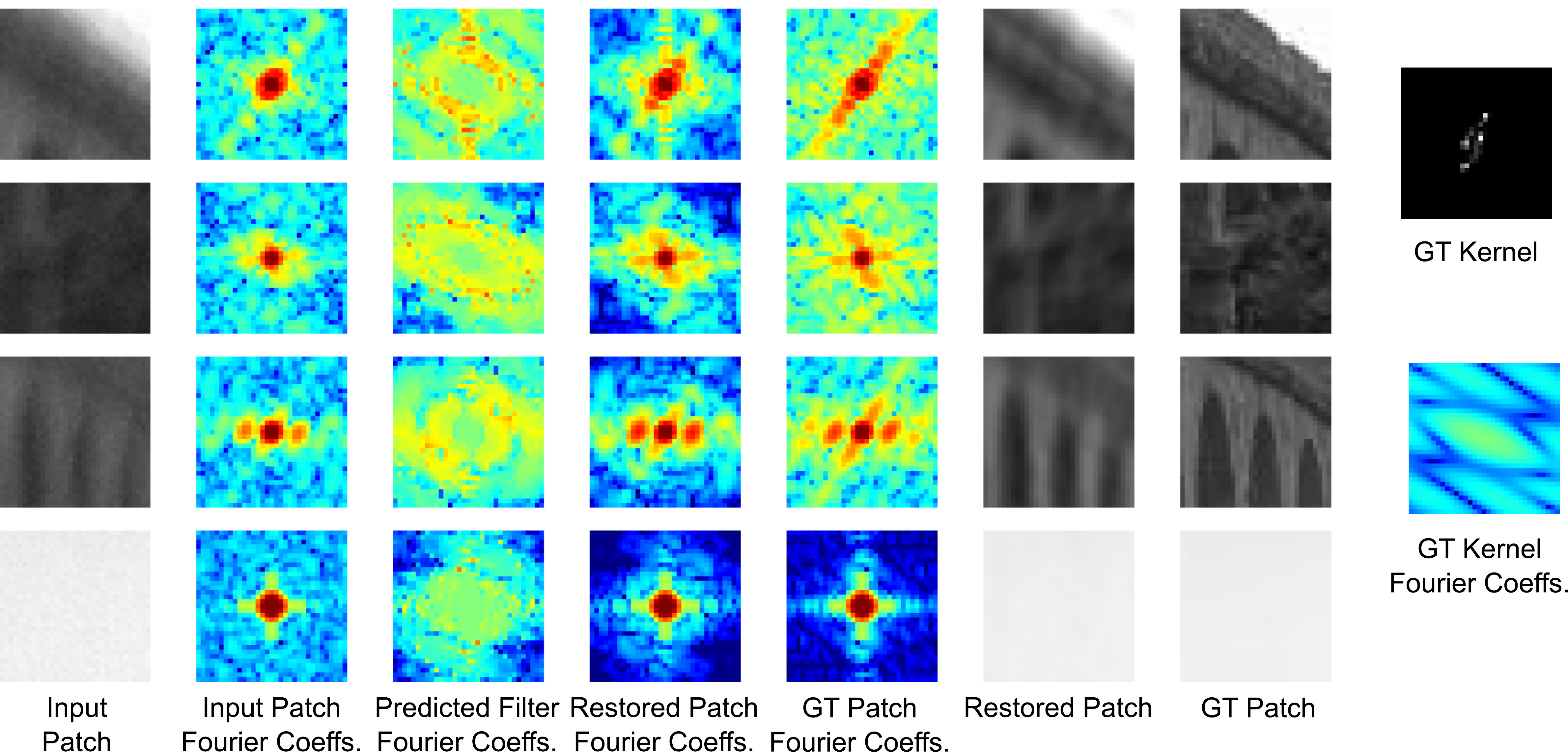}
  \caption{Examples of per-patch restoration using our network. Shown here are different patches extracted from a blurred image from \cite{sun2013}. For each patch, we show the observed blurry patch in the spatial domain, its DFT coefficients $Y[z]$ (in terms of log-magnitude), the predicted filter coefficients $G[z]$ from our network, the DFT of the resulting restored patch $X[z]$, and the restored patch in the spatial domain. As comparison, we also show the ground truth sharp image patch and its DFT, as well as the common ground truth kernel of the network and its DFT.}
  \label{fig:ffts}
\end{figure}

\subsection{Local Network Predictions}
\label{sec:lnw}

Figure~\ref{fig:ffts} illustrates the typical behavior of our trained neural network on individual patches. All patches in the figure are taken from the same image from~\cite{sun2013}, which means that they were all blurred by the same kernel (the kernel, and its Fourier coefficients, are also shown). However, we see that the predicted restoration filter coefficients are qualitatively different across these patches.

While some of this variation is due to the fact that the network is reasoning with these patches independently, remember from Sec.~\ref{sec:form} that we expect the ideal restoration filter to vary based on image content. The predicted filters in Fig.~\ref{fig:ffts} can be understood in that context as attempting to amplify different subsets of the frequencies attenuated by the blur kernel, based on which frequencies the network believes were present in the original image. Comparing the Fourier coefficients of the ground truth sharp patch to our restored outputs, we see that our network restores many frequency components attenuated in the observed patch, without amplifying noise.

These examples also validate our decision to estimate a restoration filter instead of the blur kernel from individual patches. Most patches have no content in entire ranges of frequencies even in their ground-truth sharp versions (most notably, the patch in the last row that is nearly flat), which makes estimating the corresponding frequency components of the kernel impossible. However, we are still able to restore these patches since that just requires identifying that those frequency components are absent.

Looking at the restored patches in the spatial domain, we note that while they are sharper than the input, they still have a lot of high-frequency information missing. However, remember that even our direct neural estimate $x_N[n]$ of the sharp image is composed by averaging estimates from multiple patches at each pixel (see Fig.~\ref{fig:system}, and the supplementary material for examples of these estimates). Moreover, our final estimates are computed by fitting a global kernel estimate to these locally restored outputs, benefiting from the fact that correctly restored frequencies in all patches are coherent with the same (true) blur kernel.

\subsection{Performance Evaluation}
\label{sec:perfmet}

Next, we evaluate our overall method and compare it to several recent  algorithms~\cite{irani,sun2013,xujia,cholee,levinetal,krishnan,choetal,l2d} on the Sun \etal~benchmark~\cite{sun2013}. Deblurring quality is measured in terms of the MSE between the estimated and the ground truth sharp image, ignoring a fifty pixel wide boundary on all sides in the latter, and after finding the crop of the restored estimate that aligns best with this ground truth.  Performance on the benchmark is evaluated~\cite{irani,sun2013} using quantiles of the \emph{error ratio} $r$ between the MSE of the estimated image and that of the deconvolving the observed image with the ground truth kernel using EPLL~\cite{epll}. Results with $\rgt \leq 5$ are considered to correspond to ``successful'' restoration~\cite{irani}.

\begin{figure}[!t]
  \centering
  \includegraphics[width=\textwidth]{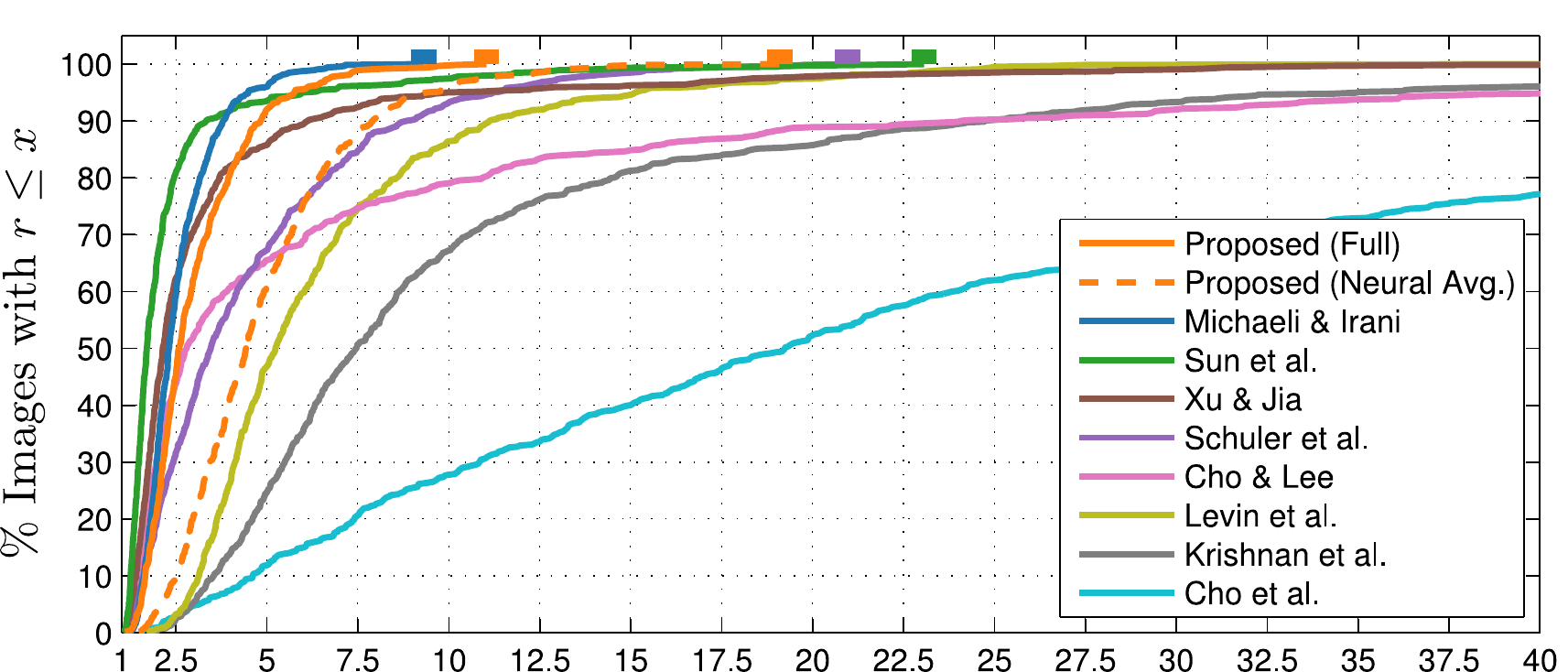}
  \caption{Cumulative distributions of the error ratio $r$ for different methods on the Sun \etal~\cite{sun2013} benchmark. These errors were computed after using EPLL~\cite{epll} for blind deconvolution using the global kernel estimates from each algorithm. The only exception is the ``neural average'' version of our method, where the errors correspond to those of our initial estimates computed by directly averaging per-patch neural network outputs, without reasoning about a global kernel.}
  \label{fig:perfmet}
\end{figure}
\begin{table}[!t]
\caption{Quantiles of error-ratio $\rgt$ and success rate ($\rgt \leq 5$), along with kernel estimation time for different methods}
\label{tab:rgt}
\centering{\small\begin{tabular}{rccccl}
  \hline
  Method & ~Mean~ & ~95\%-ile~  & ~Max~
  & ~~Success Rate~~& Time\\\hline
  (Neural Avg.) & 4.92 & 9.39 & 19.11 & 61\% &\\
  \rowcolor{tblcolor} Proposed & 3.01 & 5.76 & 11.04 & 92\% & 65s (GPU)\\
  Michaeli \& Irani~\cite{irani} & 2.57 & 4.49 & 9.31 & 96\% & 91min (CPU) \\
  Sun et al.~\cite{sun2013} & 2.38 & 5.98 & 23.07 & 93\% & 38min (CPU) \\
  Xu \& Jia~\cite{xujia} & 3.63 & 9.97 & 65.33 & 86\% & 25s (CPU) \\
  Schuler et al.~\cite{l2d} & 4.53 & 11.21 & 20.96 & 67\% & 22s (CPU) \\
  Cho \& Lee~\cite{cholee} & 8.69 & 40.59 & 111.19 & 66\% & 1s (CPU) \\
  Levin et al.~\cite{levinetal} & 6.56 & 15.13 & 40.87 & 47\% & 4min (CPU) \\
  Krishnan et al.~\cite{krishnan} & 11.65 & 34.93 & 133.21 & 25\% & 3min (CPU)\\
  Cho et al.~\cite{choetal} & 28.13 & 89.67 & 164.94 & 12\% & 1min (CPU)\\\hline
\end{tabular}
}
\end{table}

\begin{figure}[!t]
  \centering
  \includegraphics[width=\textwidth]{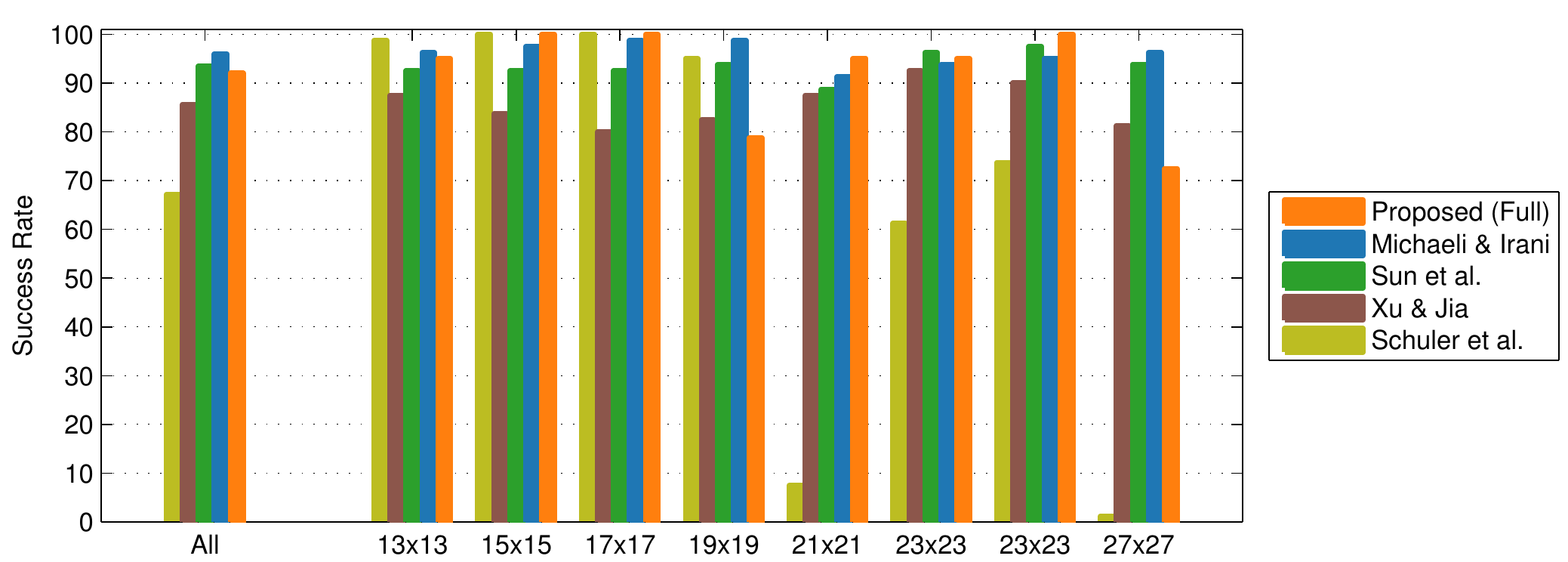}
  \caption{Success rates of different methods, evaluated over the entire Sun \etal~\cite{sun2013} dataset, and separately over images blurred with each of the 8 kernels. The kernels are sorted according to size (noted on the x-axis).}
  \label{fig:scdiff}
\end{figure}

\begin{figure}[!t]
  \centering
  \includegraphics[width=0.98\textwidth]{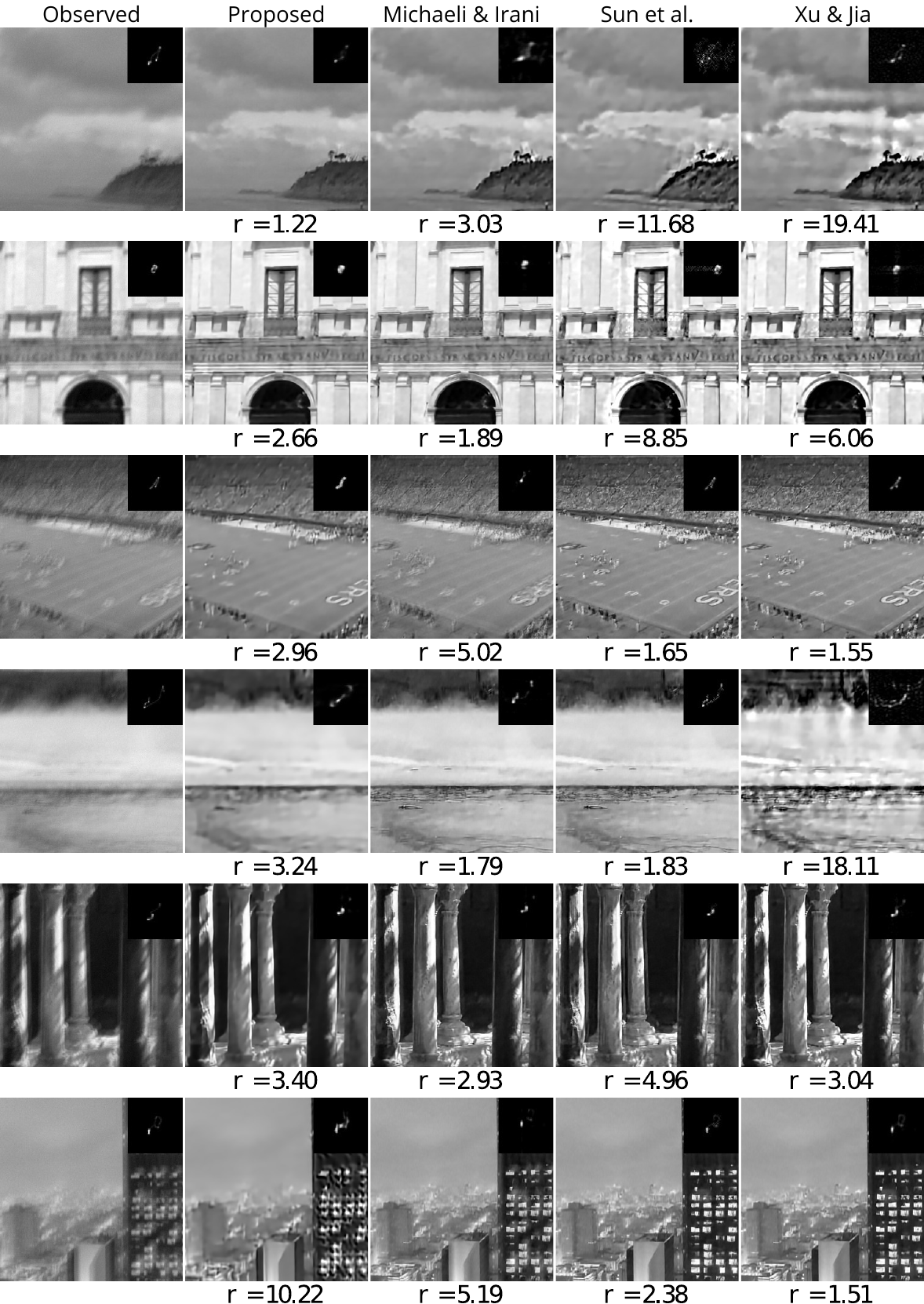}
  \caption{Example deblurred results from different methods. The estimated kernels are shown inset, with the ground truth kernel shown with the observed images in the first column. This figure is best viewed in the electronic version.}
  \label{fig:res}
\end{figure}

Figure~\ref{fig:perfmet} shows the cumulative distribution of the error-ratio for all methods on the benchmark. We also report specific quantiles of the error ratio---mean, and outlier performance in terms of $95\%$-ile and maximum value---as well as the success rate of each method in Table~\ref{tab:rgt}. The results for \cite{irani} and \cite{l2d} were provided by their authors, while those for all other methods are from \cite{sun2013}. Results for all methods were obtained using EPLL for blind-deconvolution based on their kernel estimates, and are therefore directly comparable to those of our overall method. We also report the performance of our initial estimates from just the direct neural averaging step in Fig.~\ref{fig:perfmet} and Table~\ref{tab:rgt}, which did not involve any global kernel estimation or the use of non-blind deconvolution with EPLL.

The performance of the full version of our method performs is close to that of the two state-of-the-art methods of Michaeli and Irani~\cite{irani} and Sun \etal~\cite{sun2013}. While our mean errors are higher than those of both and \cite{irani} and \cite{sun2013}, we have a near identical success rate and better outlier performance than \cite{sun2013}. Note that our method outperforms the remaining algorithms by a significant margin on all metrics. The best amongst these is the efficient approach of Xu and Jia~\cite{xujia} which is able to perform well on many individual images, but has higher errors and succeeds less often on average than our approach and that of \cite{irani,sun2013}. Figure~\ref{fig:scdiff} compares the success rate of different methods over individual kernels in the benchmark, to study the effect of kernel size. We see that the previous neural approach of \cite{l2d} suffers a sharp drop in accuracy for larger kernel sizes. In contrast, our method's performance is more consistent across the whole range of kernels in \cite{sun2013} (albeit, our worst performance \emph{is} with the largest kernel).

In addition to accuracy, Table \ref{tab:rgt} also reports the running time for kernel estimation for all methods. We see that while our method has nearly comparable performance to the two state-of-the-art methods \cite{irani,sun2013}, it offers a significant advantage over them in terms of speed. A MATLAB implementation of our method takes a total of only {65 seconds} for kernel estimation using an NVIDIA Titan X GPU. The majority of this time, 45 seconds, is taken to compute the initial neural-average estimate $x_N$. On the other hand, \cite{irani} and \cite{sun2013} take {91~minutes} and {38 minutes} respectively, using the MATLAB/C implementations of these methods provided by their authors on an I-7 3.3GHz CPU with 6 cores.

While \cite{irani,sun2013}'s running times could potentially be improved if they are reimplemented to use a GPU, their ability to benefit from parallelism is limited by the fact that both are iterative techniques whose computations are largely sequential (in fact, we only saw speed-ups of 1.4X and 3.5X in \cite{irani} and \cite{sun2013}, respectively, when going from one to six CPU cores). In contrast, our method maps naturally to parallel architectures and is able to fully saturate the available cores on a GPU. Batched forward passes through a neural network are especially efficient on GPUs, which is what the bulk of our computation involves---applying the local network \emph{independently} and in-parallel on all patches in the input image.

Some methods in Table \ref{tab:rgt} are able to use simpler heuristics or priors to achieve lower running times. But these are far less robust and have lower success rates---\cite{xujia} has the best performance amongst this set. Our method therefore provides a new and practically useful trade-off between reliability and speed.

In Fig.~\ref{fig:res}, we show some examples of estimated kernels and deblurred outputs from our method and those from \cite{irani,sun2013,xujia}. In general, we find that most of the failure cases of \cite{irani,sun2013,xujia} correspond to scenes that are a poor fit to their hand-crafted generative image priors---\eg most of \cite{sun2013,xujia}'s failure cases correspond to images that lack well-separated strong edges. Our discriminatively trained neural network derives its implicit priors automatically from the statistics of the training set, and is relatively more consistent across different scene types, with failure cases corresponding to images where the network encounters ambiguous textures that it can't generalize to. We refer the reader to the supplementary material and our project website at \url{http://www.ttic.edu/chakrabarti/ndeblur} for more results. The MATLAB implementation of our method, along with trained network weights, is also available at the latter.

\section{Conclusion}
\label{sec:conc}

In this paper, we introduced a neural network-based method for blind image deconvolution. The key component of our method was a neural network that was discriminatively trained to carry out restoration of individual blurry image patches. We used intuitions from a frequency-domain view of non-blind deconvolution to formulate the prediction task for the network and to design its architecture. For whole image restoration, we averaged the per-patch neural outputs to form an initial estimate of the sharp image, and then estimated a global blur kernel from this estimate. Our approach was found to yield comparable performance to state-of-the-art iterative blind deblurring methods, while offering significant advantages in terms of speed.

We believe that our network can serve as a building block for other applications that involve reasoning with blur. Given that it operates on local regions independently, it is likely to be useful for reasoning about spatially-varying blur---\eg arising out of defocus and subject motion. We are also interested in exploring architectures and pooling strategies that allow efficient sharing of information across patches. We expect that such sharing can be used to communicate information about a common blur kernel, to exploit ``internal'' statistics of the image (which forms the basis of the method of \cite{irani}), and also to identify and adapt to texture statistics of different scene types (\eg \cite{l2d} demonstrated improved performance when training and testing on different image categories).

\paragraph{Acknowledgments.}
The author was supported by a gift from Adobe Systems, and by the donation of a Titan X GPU from NVIDIA Corporation that was used for this research.

\end{document}